%% file: main.tex
\documentclass[journal]{IEEEtran}
\usepackage{graphicx}
\usepackage{hyperref}
\usepackage{hhline}
\usepackage{amsfonts}

\DeclareGraphicsExtensions{.pdf,.png,.jpg,.jpeg,.eps}

\ifCLASSINFOpdf
\else
\fi
\begin{document}
%

\title{NL-LinkNet: Toward Lighter but More Accurate Road Extraction with Non-Local Operations}

%
%
%

\author{Yooseung~Wang,~
        Junghoon~Seo,~and Taegyun~Jeon,~\IEEEmembership{Member,~IEEE}
\thanks{This work was done while Y.S. Wang was an intern at SI Analytics.}
\thanks{Y.S. Wang is with the Department of Electrical and Computer Engineering, Ulsan National Institute of Science and Technology, Ulsan 51144, South Korea (e-mail: yswang96@unist.ac.kr).}
\thanks{J.H. Seo is with Satrec Initiative Inc., Deajeon 34051, South Korea (email: sjh@satreci.com).}
\thanks{T.G. Jeon is with SI Analytics Inc., Deajeon 34051, South Korea (email: tgjeon@si-analytics.ai).}
}
%
%

\markboth{UNDER REVIEW ON IEEE GEOSCIENCE AND REMOTE SENSING LETTERS}
{Shell \MakeLowercase{\textit{et al.}}: Bare Demo of IEEEtran.cls for Journals}
%



\maketitle

\begin{abstract}
Road extraction from very high resolution satellite (VHR) images is one of the most important topics in the field of remote sensing. 
In this paper, we propose an efficient Non-Local LinkNet with non-local blocks that can grasp relations between global features. This enables each spatial feature point to refer to all other contextual information and results in more accurate road segmentation. 
In detail, our single model without any post-processing like CRF refinement, performed better than any other published state-of-the-art ensemble model in the official DeepGlobe Challenge. Moreover, our NL-LinkNet beat the D-LinkNet, the winner of the DeepGlobe challenge \cite{demir2018deepglobe}, with 43 \% less parameters, less giga floating-point operations per seconds (GFLOPs) and shorter training convergence time. We also present empirical analyses on the proper usages of non-local blocks for the baseline model. \end{abstract}


%
\IEEEpeerreviewmaketitle

\input{body.tex}


%




\ifCLASSOPTIONcaptionsoff
  \newpage
\fi


\input{references.tex}
\input{biograph.tex}

\end{document}

%% file: body.tex
\section{Introduction}

\IEEEPARstart{R}{oad} extraction from very high resolution (VHR) satellite images plays an important role in remote sensing applications. Valuable geographic information with high accessibility can be discovered from the extracted road network. Thus, it can be used for various tasks such as automated map update, urban planning, road navigation, unmanned vehicles, attention of road safety hazards or support to disaster relief missions. 

 Many methods on road extraction have been researched for decades. Traditional line segmentation began with the Canny edge detector \cite{Canny} and Hough transform  \cite{Hough} eventually extracting all lines, including a certain number of edge points. However, these methods contained many false detections and were time-consuming. The next step was unsupervised methods using clustering algorithms. Markov Random Field (MRF) \cite{fundamental1}  and support vector machine (SVM) \cite{fundamental03} paved the way for the automatic methods for road extraction. 

In recent years, Convolutional neural networks (CNN) have shown great improvement in detecting roads from VHR satellite images. For instance, FCN \cite{FCN, FCN02, PedNet,DenseNet} and U-Net \cite{UNet,LinkNet} involving D-LinkNet \cite{D-LinkNet}, which was the winner of the DeepGlobe 2018 Road Extraction Challenge \cite{Demir_2018_CVPR_Workshops}, achieved great improvements in scene parsing by introducing end-to-end learning. Nonetheless, small receptive fields of CNNs have been considered as a challenge.
To solve it, the dilated convolutions \cite{deeplabv3, deeplabv3+,Deformable} are fairly improved with the larger receptive field by proposing the novel method of extending kernel size of the convolution. In parallel, Zhao \emph{et al.} \cite{PSA,PSA2,PSA3} proposed the point-wise spatial attention network for expanding receptive fields.


Our major concern is that even the dilated convolution had limitations when capturing long-range dependencies. Since satellite imagery is shot overhead, local features might be covered by other obstacles such as shadows, clouds, trees or buildings while the single convolution layer considers only a few neighborhoods depending on its kernel size. Moreover, deeper CNN architecture is well known not to guarantee a larger effective receptive field \cite{luo2016understanding}.

In this paper, we propose our novel network named \textbf{N}on-\textbf{L}ocal \textbf{LinkNet} (\textbf{NL-LinkNet}) with differentiable non-local operations in order to tackle this problem. Our perspective is directly inspired by very recent works on non-local neural network \cite{NonLocal2018} and its image restoration applications \cite{resnonlocal}. 
  Non-local neural operations compute feature map values as a weighted sum of the features at all positions. Thus, it enables the model to capture distant information and to account for long-range dependencies efficiently. 

In sum, the main contributions of this paper are the threefold.
\begin{enumerate}
\item  our \emph{single} model \emph{without} any sophiscated post-processing like conditional random field (CRF) showed higher performance than any previously published \emph{ensemble} model in the official DeepGlobe Challenge. NL-LinkNet even outperformed the 1st ranked solution \cite{D-LinkNet} in the official DeepGlobe Challenge \cite{demir2018deepglobe} using 43\% less parameters with lower GFLOPs and shorter training convergence time. 

\item 
We represent empirical analyses of non-local operations to extract the roads. NL-LinkNet achieved more promising results than the baseline (without any non-local blocks) for all combinations of distinct pairwise functions and locations of non-local blocks. Both single and double non-local block models showed superior performance to the baseline.

\item We did the first work to use neural non-local operations for road extraction on VHR satellite images. Non-local operations enable the model to capture long-range dependencies to solve geographic limitations of roads partly covered by objects such as shadows, clouds, buildings or trees.

\end{enumerate}


\begin{figure}
  \centering
    \includegraphics[width=0.37\textwidth]{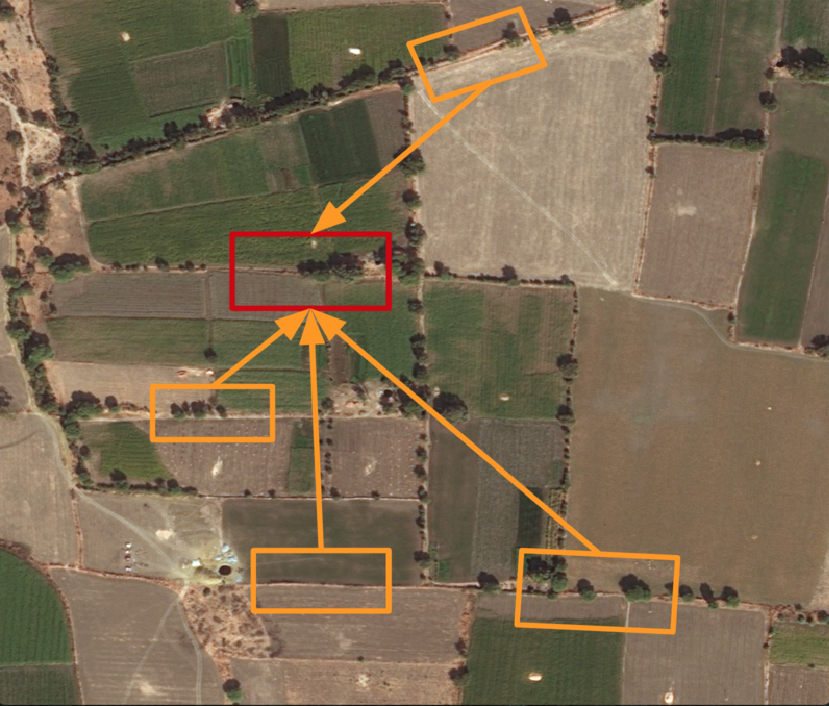} {}
  \caption{Illustration of non-local operation for road extraction from a VHR satellite image. In order to detect the road in the red box blocked by trees, non-local operation needs to refer to other roads in orange boxes.}
\label{non-local-illust}
\end{figure}

\section{Non-local Network for road extraction}
\subsection{Non-local operations for road extraction}
Since satellite images are taken overhead at high altitude, they are likely to be blocked by other obstacles. Roads themselves are the most likely to be blocked by other obstacles such as the shadows of tall buildings and trees covering roads. To overcome this situation, capturing long-range dependencies is essential, but almost CNN methods have difficulty solving long-range dependency. Convolution operation refers to only local information through a small kernel, which makes it difficult to refer to distant information. Recurrent operation, another local method, is based only on current and previous features. Local methods construct these  operations repeatedly encountering many drawbacks such as lower efficiency of memory and computation for the optimization of the model.

Therefore, we borrowed the concept of the non-local operation \cite{NonLocal2018} model's long-range dependencies. A non-local operation computes a feature map as a weighted sum of all pixels, and it allows the model to capture long-range dependencies and distant information. In Fig. \ref{non-local-illust}, the road in the middle (red box) is covered by trees. Local methods cannot recognize the relationships between relevant roads. Non-local networks, however, refer to (orange boxes) near roads and extract the covered roads correctly. More examples are shown in Fig. \ref{visual-examples}.

\begin{figure}
  \centering
    \includegraphics[width=0.46\textwidth]{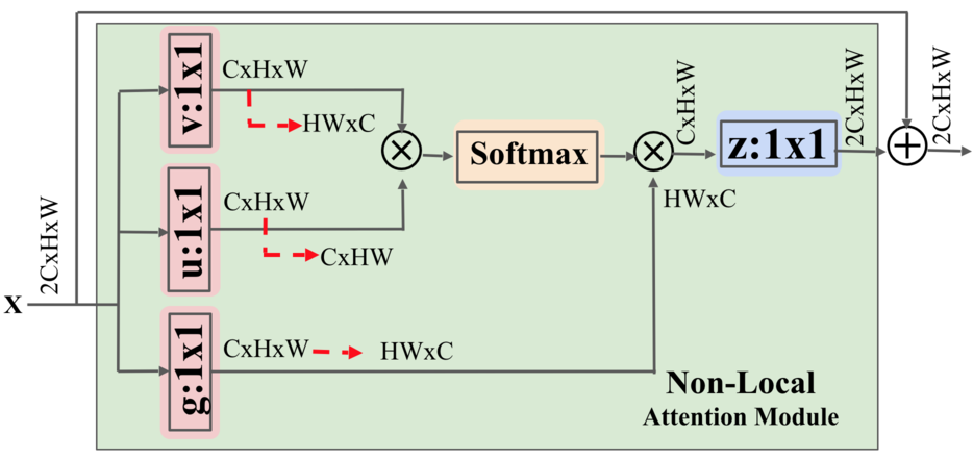}{}
  \caption{A non-local block with the embedded Gaussian $f$. The Gaussian version $f$ has no $u$, $v$ function. The dot-product version $f$ has no softmax function and divides output results by $N$. $H$, $W$, and $C$ (height, width, and channel, respectively). $\oplus$ denotes element-wise sum and $\otimes$ denotes matrix multiplication. The green colored part represents the non-local attention module.}
\label{non-local-block}
\end{figure}

\begin{figure*}[ht]
  \centering
    \label{fig_first_case}
    \includegraphics[width=0.75\textwidth]{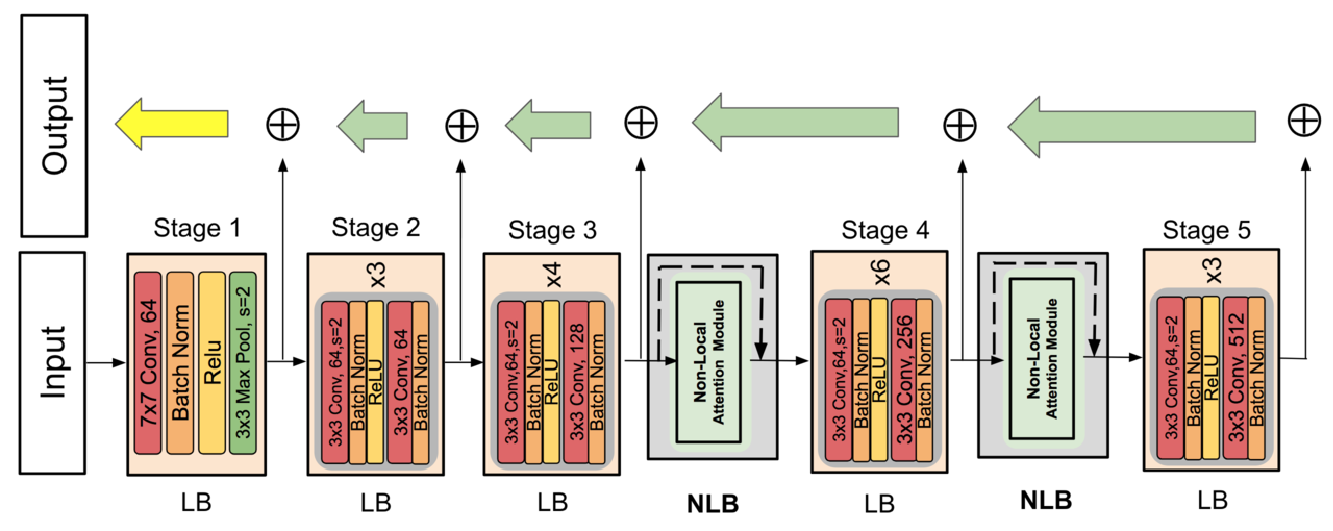} {}
  \caption{An architecture of our NL-LinkNet for road extraction. \emph{LB} is local block and \emph{NLB} is non-local block. \emph{Green arrows} show the consequences of 1x1 convolution, 3x3 transpose convolution with stride 2, and 1x1 convolution. The \emph{Yellow arrow} refers to 4x4 transpose convolution and 3x3 convolution followed by the green arrow. The details of the NLB with non-local attention module are shown in Fig. \ref{non-local-block}.}
\label{our-nln}
\end{figure*}

\subsection{Formulation} 
We used generic non-local operations in non-local networks \cite{NonLocal2018} for road extraction: 
\begin{equation} 
\mathbf{y}_{i} = \frac{1}{\mathbf{C}}\sum_{j=1}^N f(\mathbf{x}_{i},\mathbf{x}_{j})g(\mathbf{x}_{j}), \label{eq:1}
\end{equation} 
where $N$ is the input dimension, $i \in \mathbb{N}$ is the index of an output feature map, and $j \in \mathbb{N}$ is the index of all possible positions. $\mathbf{x} \in \mathbb{R}^{N \times c_{1}}$ is the input feature map and $\mathbf{y} \in \mathbb{R}^{N \times c_{2}}$ is the output feature map where $c_{1},c_{2}$ are the number of input channels and output channels.  $\mathbf{x}_{i}, \mathbf{x}_{j} \in \mathbb{R}^{c_{1}}$ and $\mathbf{y}_{i} \in \mathbb{R}^{c_{2}}$ indicate the $i$-th and $j$-th feature of $\mathbf{x}$ and the $i$-th feature of $\mathbf{y}$, respectively. A pairwise function $f : \mathbb{R}^{c_{1}}\times\mathbb{R}^{c_{1}}\rightarrow\mathbb{R} $ computes a scalar that reflects the correlation between $\mathbf{x}_{i}$ and $\mathbf{x}_{j}$. The unary function $g : \mathbb{R}^{c_{1}} \rightarrow \mathbb{R}^{c_{2}}$ represents the input signal $\mathbf{x}_{j}$ as the $c_{2}$ channel.
Finally, the response is normalized by a factor $\mathbf{C} \in \mathbb{R}^{+}$. Compared with traditional convolution or recurrent operations, the non-local operation (\ref{eq:1}) computes the weighted average of all the features.
\subsection{Instantiation of Non-local Operations and Non-local Block}
There are several possible choices for $f$ and $g$. We chose a simple form of linear embedding for $g$; $g(\mathbf{x}_{j}) = W_{g}\mathbf{x}_{j}$. The parameter matrix to be learned is $W_{g} \in \mathbb{R}^{c_{2} \times c_{1}}$
For the pairwise function $f$, we considered three pairwise functions.

First, we simply used the dot product to evaluate the pairwise relationship:

\begin{equation}
f(\mathbf{x}_{i},\mathbf{x}_{j}) =  u(\mathbf{x}_{i})^{T}v(\mathbf{x}_{j}) = (W_{u}\mathbf{x}_{i})^{T}(W_{v}\mathbf{x}_{j}), \label{eq:2}
\end{equation}
where $W_{u}, W_{v} \in \mathbb{R}^{c_{2} \times c_{1}}$ are weight matrices; $u, v : \mathbb{R}^{c_{1}} \rightarrow \mathbb{R}^{c_{2}}$. We used $C = N$ for the dot product pairwise function. Next, we considered embedded Gaussian version (\ref{eq:4}) and Gaussian version (\ref{eq:5}):
\begin{equation}
f(\mathbf{x}_{i},\mathbf{x}_{j}) =  e^{u(\mathbf{x}_{i})^{T}v(\mathbf{x}_{j})}
= e^{(W_{u}\mathbf{x}_{i})^{T}(W_{v}\mathbf{x}_{j})} \label{eq:4}
\end{equation} 
\begin{equation}
f(\mathbf{x}_{i},\mathbf{x}_{j}) = e^{\mathbf{x}_{i}^{T}\mathbf{x}_{j}} \label{eq:5}
\end{equation}
For both (\ref{eq:4}) and (\ref{eq:5}), we chose $C = \sum_{j=1}^{N}f(\mathbf{x}_{i},\mathbf{x}_{j})$ for these pairwise functions. In order to simplify implementation, we used dot-product similarity for the Gaussian version of $f$.

Lastly, we defined the non-local block by connecting the result of the non-local operation with an input feature via residual connection:
\begin{equation} 
\mathbf{z}_{i} = W_{z}\mathbf{y}_{i} + \mathbf{x}_{i} \label{eq:6}
\end{equation}
where $\mathbf{y}$ is given in (\ref{eq:1}) and $W_{z} \in \mathbb{R}^{c_{1} \times c_{2}}$ are weight matrices. The residual connection in the final step allows us to insert the non-local block into any position of pre-trained networks by initializing the weight matrix as zero. Figure \ref{non-local-block} illustrates an example of a non-local block with a Gaussian pairwise function. Since satellite images are two dimensional images, we employed $1\times1$ convolutions for all weight matrices $W_{u}, W_{v}, W_{g}, W_{z}$, (instead of $1\times1\times1$ convolutions). In addition, half of the channels in $\mathbf{x}$ were used for the internal channel; \emph{i.e.} $c_2 = c_1/2$. This enabled the model to reduce computation and increase the number of non-local blocks in the whole architecture.

\begin{figure}[t]
    \centering
     \includegraphics[width=0.49\textwidth]{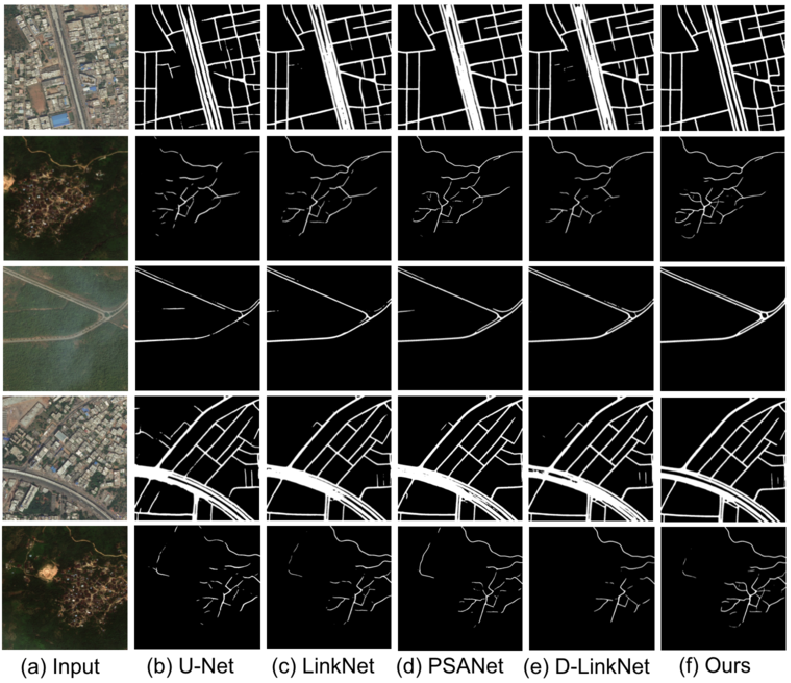} {}
    \caption{Qualitative comparisons on results for the validation set of DeepGlobe 2018 Road Extraction Challenge dataset. (a) Input image. (b) U-Net \cite{UNet}. (c) LinkNet \cite{LinkNet}. (d) PSANet \cite{PSA} (e) D-LinkNet \cite{D-LinkNet}. (f) Our proposed NL-LinkNet. Note that the PSANet represents the LinkNet with Point-wise Spatial Attention (PSA) Block at the end of the encoder.}
\label{visual-examples}
\end{figure}

\subsection{Non-local Network Architectures}
\label{section-NLNet}
 The entire architecture of our NL-LinkNet is illustrated in Fig. \ref{our-nln}. For the decoder part, we used the decoder block (green arrow) inspired by \cite{LinkNet}. We construct our decoder architectures with the combination of local block(LB) and non-local block(NLB). The local blocks refer to backbone network using the traditional local method and we employed ResNet34 \cite{resNet}. In other words, the LBs in the encoder consist of layers in each of the 5 stages from ResNet34 for the rest of experiments. While D-LinkNet, the winner in the challenge \cite{Demir_2018_CVPR_Workshops}, employed a dilated block(DB) at the end of the encoder part, our methods adopted the NLB using non local operations to achieve better efficiency in road extraction as shown in Table \ref{benchmark} and \ref{different-loc}. When designing the architecture, we considered the following two factors:
 
\paragraph{Location of non-local blocks}  To maximize efficiency of the non-local block, we considered the following to choose their location in our network. 
In order to increase the utilities of non-local blocks in various settings, efficient usage of the memory is crucial for searching the distinct effects of non-local operations. Therefore, non-local blocks after the 3rd and 4th layer were employed in our experimental settings. The results and analysis are described in \ref{location-block-section} and Table \ref{different-loc}.

\paragraph{Choice of local blocks}
The reason why we used layers in each stage from ResNet34 as our local block for the rest of the experiments is because recent state-of-the-art models \cite{LinkNet, D-LinkNet} have shown that small backbone networks such as ResNet18 and ResNet34 are effective for road extraction. Furthermore, since the relatively small ResNet34 has room for memory, we could experiment on non-local blocks with various settings.

\section{Experiments} 
\subsection{Dataset and Evaluation Metric}
In order to check the effectiveness for road extraction, we used the DeepGlobe 2018 Road Extraction Challenge dataset \cite{Demir_2018_CVPR_Workshops} for our experiments.
The width and height of all the images in the dataset are 1024, 1024. Each image is an RGB image with a 0.5 ground sampling distance (GSD), collected by DigitalGlobe's satellite.
The mask is a gray scaled binary image of the same height and width as the input image.
In the mask image, the white stands for roads and the black stands for the backgrounds.
The dataset consists of 8,579 images in total, including 6,226 images for training, 1,243 for validation, and 1,110 for testing.
We hired mean intersection-over-union (mIOU) for the evalutation metric. mIOU is the same evaluation metric used in DeepGlobe 2018 Road Extraction Challenge. To evaluate the models accurately, the labeling for roads and non-roads pixels are classified to 255 or 0, repectively. 

\subsection{Implementation details}
We employed fine-tuned model weights pretrained on ImageNet. Using same settings with \cite{LinkNet, D-LinkNet}, the 1000-way classifier in the last layer was replaced by the decoder with the sigmoid function to use the binary cross entropy loss function. 
  
\paragraph{Training Phase} 
We used binary cross entropy + dice coefficient loss as the loss function and optimized the model with the Adam optimizer \cite{kingma2014adam}. We trained the model with a batch size of 8 on 4 Titan 1080 GPUs with 12GB on-board memory and the crop size was $1024\times1024$. We did a grid search of the learning rate in the range between 1e-4 and 1e-3 with the interval of 1e-5. The learning rate starting from 0.0003 fell off step by step by 0.2 when loss did not fall. The network converged in 180 epochs and the convergence took 47 hours. Some data augmentation techniques were done including image shifting, multi-scaling, vertical flip and horizontal flip for each input. 

\paragraph{Test Phase}
Test time augmentation was done including image horizontal flip, vertical flip, and diagonal flip. For some experiments, we did multi-scale testing (MS) with scales of [0.75, 1.0, 1.25].
Note that we did not use more sophisticated post-processing methods like CRF. Without using CRF, our models outperformed all the other published methods.

\setlength{\doublerulesep}{2pt}

\section{Results and Discussions}
\subsection{Performance Evaluation}
\label{quantitative}\textbf{}
The baseline in Table \ref{different-loc} and \ref{different-f} refer the architecture in Fig \ref{our-nln} without any NLB, LB, or multi-scaling for the initial setting. In Table \ref{different-loc}, PSAB, DB, NLB3 and NLB4 are blocks for PSA~\cite{PSA}, Dilated~\cite{D-LinkNet}, Non-local~\cite{Classic_nonlocal} (at stage 3 and 4) operations. NL3-LinkNet, NL4-LinkNet and NL34-LinkNet use NLB3 and NLB4, adding the NLB at either the 3rd, 4th, or both stages at the encoder.
PSANet and D-LinkNet employee the DB and PSAB, at the end of encoder.

\subsubsection{Benchmarks Comparison} 
Table \ref{benchmark} describes the results of different single model methods on the DeepGlobe 2018 Road Extraction Challenge validation dataset. As shown in the table, 
our NL-LinkNet achieved the best performance (65.00\%) among all the other methods \emph{without} any sophisticated post processing such as CRF. 
Further, our \emph{single} model outperformed even the ensemble model of D-LinkNet \cite{D-LinkNet}, U-Net \cite{UNet}, and LinkNet \cite{LinkNet}, which recorded 64.66\%
and won the first place.
For a detailed description of NL-LinkNet in Table \ref{benchmark}, please refer section \ref{location-block-section}. 

\begin{table}[t]
\centering
\caption{Benchmarks for DeepGlobe Road Extraction Challenge}
\label{my-label}
\resizebox{.40\textwidth}{!}{%
\begin{tabular}{ccc}
\hline
\multicolumn{1}{|c|}{BenchMarks}       & \multicolumn{1}{c|}{mIOUs} & \multicolumn{1}{c|}{Remarks}      \\ \hhline{|=|=|=|}
\multicolumn{1}{|c|}{EosResUNet \cite{EOSResUNet}}       & \multicolumn{1}{c|}{55.96} & \multicolumn{1}{c|}{}             \\
\multicolumn{1}{|c|}{StackedUNet \cite{StackedUNet}}      & \multicolumn{1}{c|}{60.60} & \multicolumn{1}{c|}{4th place}    \\
\multicolumn{1}{|c|}{ResInceptSkipNet \cite{ResInceptSkipNet}} & \multicolumn{1}{c|}{61.30} & \multicolumn{1}{c|}{3rd place}    \\
\multicolumn{1}{|c|}{U-Net \cite{UNet}}     & \multicolumn{1}{c|}{62.94} & \multicolumn{1}{c|}{from \cite{D-LinkNet}} \\
\multicolumn{1}{|c|}{LinkNet \cite{LinkNet}}      & \multicolumn{1}{c|}{63.00} & \multicolumn{1}{c|}{}             \\
\multicolumn{1}{|c|}{PSANet \cite{PSA}}              & \multicolumn{1}{c|}{63.87} & \multicolumn{1}{c|}{}    \\
\multicolumn{1}{|c|}{FCN \cite{FCN06}}              & \multicolumn{1}{c|}{64.00} & \multicolumn{1}{c|}{2nd place}    \\
\multicolumn{1}{|c|}{D-LinkNet \cite{D-LinkNet}}        & \multicolumn{1}{c|}{64.12} & \multicolumn{1}{c|}{1st place}    \\
\multicolumn{1}{|c|}{\cite{UNet}+\cite{LinkNet}+\cite{D-LinkNet}}        & \multicolumn{1}{c|}{64.66} & \multicolumn{1}{c|}{from repo. of \cite{D-LinkNet}}    \\
\hhline{|=|=|=|}

\multicolumn{1}{|c|}{\textbf{NL-LinkNet}}      & \multicolumn{1}{c|}{\textbf{65.00}} & \multicolumn{1}{c|}{\textbf{Ours}}         \\ \hline 
                                       &                            &                                  
\end{tabular}%
}
\label{benchmark}
\end{table}

\subsubsection{The efficiency of NL-LinkNet} 
\label{efficiency_NLLinkNet}
The NL-LinkNet surpassed the first-ranked solution \cite{D-LinkNet} not only in accuracy, but also in efficiency. NL-LinkNet has higher accuracy, than D-LinkNet while our method has only 67\% parameters and less GFLOPs as shown in Table \ref{benchmark} and \ref{different-loc}. This implies improved efficiency of lighter non-local blocks for road extraction than the dilated blocks in D-LinkNet. 
Compared to PSANet, which also takes into account the values of all pixels without the function of g, our model with non-local blocks showed 0.72\% improved mIOU while  requiring 31\% less parameters and less GFLOPs. These results prove that non-local operations outperform the point-wise spatial module, which is its main contribution of the PSANet \cite{PSA}.

\begin{figure}
  \centering
    \includegraphics[width=0.35\textwidth]{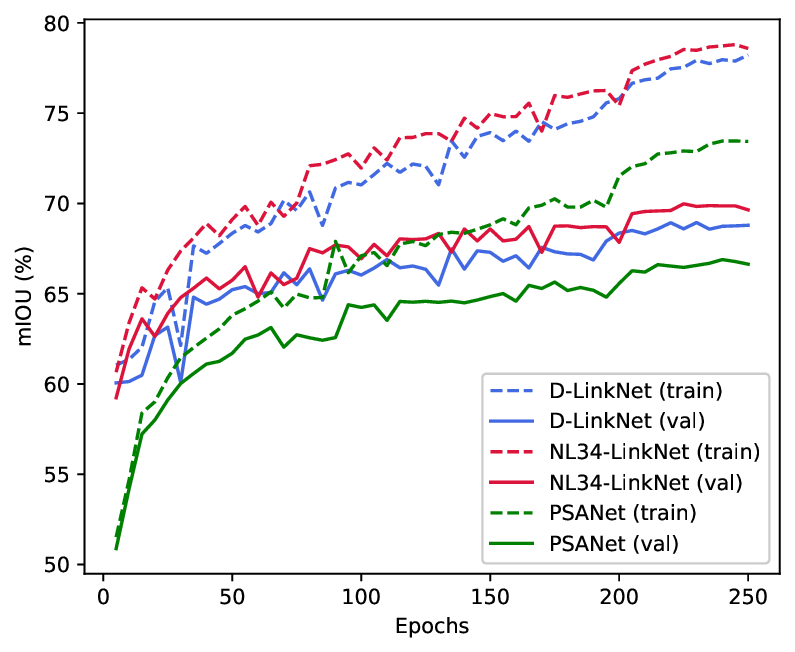} {}
  \caption{Curves of the training procedure on Road Extraction dataset in DeepGlobe Challenge for NL34-LinkNet, PSANet and D-LinkNet. Because DeepGlobe Road dataset does not have ground truth label on its own validation set, we split the original 6,226 training data into 5,026 training data and 1,200 validation data. We show the training error and validation error.}
\label{curve-of-training-procedure}
\end{figure}

  Figure \ref{curve-of-training-procedure} shows the mIOU of outputs of the D-LinkNet, PSANet and NL34-LinkNet on the training and validation set of DeepGlobe Challenge. The NL-LinkNet guarantees faster convergence speed and higher performance than D-LinkNet and PSANet on the DeepGlobe Challenge Dataset. Note that even the mIOUs in the early training phases represent significantly the better performance of the NL-LinkNet. 

\begin{table}[]
\centering
\caption{Performances of blocks added into different stages}
\label{different-loc}
\resizebox{.45\textwidth}{!}{%
\begin{tabular}{|c|cccc|c|c|c|}
\hline
Models & 
\multicolumn{1}{l}{PSAB} &
\multicolumn{1}{l}{DB} & 
\multicolumn{1}{l}{NLB3} & \multicolumn{1}{l|}{NLB4} & 
mIOUs & Params (M) & GFLOPs (B) \\ \hhline{|=|====|=|=|=|}
Baseline & X & X & X & X & 63.07 & 21.66 & 124.79 \\
PSANet & O & X & X & X &63.87 & 28.47 & 131.54 \\
D-LinkNet & X & O & X & X &64.01 & 31.10 & 134.45 \\
NL3-LinkNet & X & X & O & X & 64.15 & \textbf{21.69} & 125.34 \\
NL4-LinkNet & X & X & X & O & 64.40 & 21.79 & \textbf{125.33} \\ 
NL34-LinkNet & X & X & O & O & \textbf{64.59} & 21.82 & 124.88 \\ \hline
\end{tabular}%
}
\end{table}

\subsection{Visual Comparison}
Figure \ref{visual-examples} shows the visual performance of different methods on road extraction. The first, third and fourth row represent the case of the complex urban area with a wide central road. The second and fifth row of the figure substantiates the superiority of our method in the case of a winding region and roundabout. 
This implies that non-local operation can help capture related information with the maximized size of the receptive field. 
Meanwhile, other previous traditional local methods using small receptive fields could not detect roads in complicated cases. Even D-LinkNet with dilated convolution having an increased size of receptive field, cannot capture it clearly. 

\subsection{Analyses of different settings}
\label{subsection-diff-settings}
\subsubsection{On location of non-local blocks}
\label{location-block-section}
Table \ref{different-loc} illustrates the performance of NL-LinkNet with non-local blocks at different locations. 
We use ResNet34 layers as our local blocks and the embedded Gaussian version of $f$. 

Note that NL-LinkNet outperforms the baseline in all three combinations (NL3-LinkNet, NL4-LinkNet, NL34-LinkNet). The table indicates that adding non-local blocks achieves at least a 1.08\% improvement and adding only one non-local block (NL4-LinkNet) leads to a 1.33\% improvement over the baseline. This result proves high usability and usefulness of non-local operation at various locations. It also implies that the maximized receptive field of the non-local block can link all the condensed information for better understanding of contextual information.

\begin{table}[t]
\centering
\caption{Performances on different pairwise functions}
\label{different-f}
\resizebox{.23\textwidth}{!}{%
\begin{tabular}{|c|c|}
\hline
Models     & mIOUs          \\ \hhline{|=|=|}
Baseline       & 63.07          \\
Dot-product    & 64.04            \\ 
Gaussian       & 64.23          \\
Embedded Gaussian & \textbf{64.59} \\ \hline

\end{tabular}%
}
\end{table}

\subsubsection{On different pairwise functions} Table \ref{different-f} compares performances depending on different types of pairwise functions. Except for the baseline, NL34-LinkNet was used in the experiment. Embedded Gaussian and Gaussian version $f$ showed noteworthy improvements on mIOU. On the contrary, the dot-product version of $f$ achieved relatively slight improvement on performance. With an embedded Gaussian pairwise function, adding two non-local blocks led to a 1.52\% improvement over the baseline.

\section{Conclusions}
In this paper, we proposed the Non-Local LinkNet for road extraction from VHR satellite images. The non-local operation captures distant information referencing all the features in the satellite images. Our NL-LinkNet achieved 65.00\% mIOU scores: higher than  any  of  the  other  publications  in  the official DeepGlobe 2018 Road Extraction Challenge. It also outperformed the 1st ranked solution with less parameters, less GFLOPs, and faster training convergence time. We explored non-local blocks by varying locations and pairwise functions.

%% file: references.tex

\bibliographystyle{IEEEtran}
\bibliography{IEEEabrv,bibtex}

%% file: biograph.tex
%








%% file: main.bbl
\begin{thebibliography}{10}
\providecommand{\url}[1]{#1}
\csname url@samestyle\endcsname
\providecommand{\newblock}{\relax}
\providecommand{\bibinfo}[2]{#2}
\providecommand{\BIBentrySTDinterwordspacing}{\spaceskip=0pt\relax}
\providecommand{\BIBentryALTinterwordstretchfactor}{4}
\providecommand{\BIBentryALTinterwordspacing}{\spaceskip=\fontdimen2\font plus
\BIBentryALTinterwordstretchfactor\fontdimen3\font minus
  \fontdimen4\font\relax}
\providecommand{\BIBforeignlanguage}[2]{{%
\expandafter\ifx\csname l@#1\endcsname\relax
\typeout{** WARNING: IEEEtran.bst: No hyphenation pattern has been}%
\typeout{** loaded for the language `#1'. Using the pattern for}%
\typeout{** the default language instead.}%
\else
\language=\csname l@#1\endcsname
\fi
#2}}
\providecommand{\BIBdecl}{\relax}
\BIBdecl

\bibitem{demir2018deepglobe}
I.~Demir, K.~Koperski, D.~Lindenbaum, G.~Pang, J.~Huang, S.~Basu, F.~Hughes,
  D.~Tuia, and R.~Raska, ``Deepglobe 2018: A challenge to parse the earth
  through satellite images,'' in \emph{Proc. CVPR Workshops}, 2018.

\bibitem{Canny}
J.~Canny, ``A computational approach to edge detection,'' \emph{Pattern
  Analysis and Machine Intelligence, IEEE Transactions on}, vol. PAMI-8, pp.
  679 -- 698, 12 1986.

\bibitem{Hough}
R.~O. Duda and P.~E. Hart, ``Use of the hough transformation to detect lines
  and curves in pictures.'' \emph{Commun. ACM}, vol.~15, no.~1, pp. 11--15,
  1972.

\bibitem{fundamental1}
F.~Tupin, H.~Maitre, J.-F. Margin, J.-M. Nicolas, and E.~Pechersky, ``Detection
  of linear features in sar images: Application to road network extraction,''
  \emph{IEEE Trans. Geosci. Remote Sense.}, 1998.

\bibitem{fundamental03}
F.~Hu, G.-S. Xia, J.~Hu, and L.~Zhang, ``Road extraction using svm and image
  segmentation,'' \emph{Photogramm. Eng. Remote Sens.}, vol.~70, no.~12, pp.
  1365--1371, 2004.

\bibitem{FCN}
J.~Long, E.~Shelhamer, and T.~Darrell, ``Fully convolutional networks for
  semantic segmentation,'' in \emph{Proc. CVPR}, 2015.

\bibitem{FCN02}
Y.~Wei, Z.~Wang, and M.~Xu, ``Road structure refined cnn for road extraction in
  aerial image,'' \emph{IEEE Geosci. Remote Sens. Letters}, 2017.

\bibitem{PedNet}
M.~Ullah, A.~Mohammed, and F.~Alaya~Cheikh, ``Pednet: A spatio-temporal deep
  convolutional neural network for pedestrian segmentation,'' \emph{Journal of
  Imaging}, vol.~4, p. 107, 09 2018.

\bibitem{DenseNet}
S.~Jegou, M.~Drozdzal, D.~Vazquez, A.~Romero, and Y.~Bengio, ``The one hundred
  layers tiramisu: Fully convolutional densenets for semantic segmentation,''
  \emph{Proc. CVPR Workshops}, 2017.

\bibitem{UNet}
Z.~Zhang, Q.~Liu, and Y.~Wang, ``Road extraction by deep residual u-net,''
  \emph{IEEE Geosci. Remote Sens. Letters}, May. 2015.

\bibitem{LinkNet}
L.~Zhou, C.~Zhang, and M.~Wu, ``Linknet: Exploiting encoder representations for
  efficient semantic segmentation,'' in \emph{Proc. VCIP}, 2017.

\bibitem{D-LinkNet}
L.~Zhou, C.~Zhang, and W.~M, ``D-linknet: Linknet with pretrained encoder and
  dilated convolution for high resolution satellite imagery road extraction,''
  in \emph{Proc. CVPR Workshops}, 2018.

\bibitem{Demir_2018_CVPR_Workshops}
I.~Demir, K.~Koperski, D.~Lindenbaum, G.~Pang, J.~Huang, S.~Basu, F.~Hughes,
  D.~Tuia, and R.~Raskar, ``Deepglobe 2018: A challenge to parse the earth
  through satellite images,'' in \emph{Proc. CVPR Workshops}, 2018.

\bibitem{deeplabv3}
L.-C. Chen, G.~Papandreou, F.~Schroff, and H.~Adam, ``Rethinking atrous
  convolution for semantic image segmentation,'' \emph{arXiv preprint
  arXiv:1706.05587}, 2017.

\bibitem{deeplabv3+}
L.-C. Chen, Y.~Zhu, G.~Papandreou, F.~Schroff, and H.~Adam, ``Encoder-decoder
  with atrous separable convolution for semantic image segmentation,'' in
  \emph{Proc. ECCV}, 2018.

\bibitem{Deformable}
J.~Dai, H.~Qi, Y.~Xiong, Y.~Li, G.~Zhang, H.~Hu, and Y.~Wei, ``Deformable
  convolutional networks,'' \emph{arXiv preprint arXiv:1703.06211}, 2017.

\bibitem{PSA}
H.~Zhao, S.~Liu, S.~Jianping, C.~C. Loy, D.~Lin, and J.~Jia, ``Psanet:
  Point-wise spatial attention network for scene parsing,'' in \emph{Proc.
  ECCV}, 2018.

\bibitem{PSA2}
H.~Wang, Y.~Fan, Z.~Wang, L.~Jiao, and B.~Schiele, ``Parameter-free spatial
  attention network for person re-identification,'' \emph{arXiv preprint
  arXiv:1811.12150}, 2018.

\bibitem{PSA3}
C.~Han, F.~Shen, L.~Liu, Y.~Yang, and H.~T. Shen, ``Visual spatial attention
  network for relationship detection,'' in \emph{Proc. MM}, 2018.

\bibitem{luo2016understanding}
W.~Luo, Y.~Li, R.~Urtasun, and R.~Zemel, ``Understanding the effective
  receptive field in deep convolutional neural networks,'' in \emph{Proc.
  NIPS}, 2016.

\bibitem{NonLocal2018}
X.~Wang, R.~Girshick, A.~Gupta, and K.~He, ``Non-local neural networks,'' in
  \emph{Proc. CVPR}, 2018.

\bibitem{resnonlocal}
Y.~Zhang, K.~Li, K.~Li, Z.~Bineng, and Y.~Fu, ``Residual non-local attention
  networks for image restoration,'' in \emph{Proc. ICLR}, 2019.

\bibitem{resNet}
K.~He, X.~Zhang, S.~Ren, and J.~Sun, ``Deep residual learning for image
  recognition,'' in \emph{Proc. CVPR}, 2016, pp. 770--778.

\bibitem{kingma2014adam}
D.~P. Kingma and J.~Ba, ``Adam: A method for stochastic optimization,'' in
  \emph{Proc. ICLR}, 2015.

\bibitem{Classic_nonlocal}
A.~Buades, B.~Coll, and J.-M. Morel, ``A non-local algorithm for image
  denoising,'' in \emph{Proc. CVPR}, no.~5, 2005, pp. 709--713.

\bibitem{EOSResUNet}
O.~Filin, A.~Zapara, and S.~Panchenko, ``Road detection with eosresunet and
  post vectorizing algorithm,'' in \emph{Proc. CVPR Workshops}, 2018.

\bibitem{StackedUNet}
T.~Sun, Y.~Wenxiang, and Y.~Wang, ``Stacked u-nets with multi-output for road
  extraction,'' in \emph{Proc. CVPR Workshops}, 2018.

\bibitem{ResInceptSkipNet}
J.~Doshi, ``Residual inception skip network for binary segmentation,'' in
  \emph{Proc. CVPR Workshops}, 2018.

\bibitem{FCN06}
Z.~Zhong, J.~Li, W.~Cui, and H.~Jiang, ``Fully convolutional networks for
  building and road extraction: Preliminary results,'' in \emph{Proc. IEEE
  IGARSS}, 2016.

\end{thebibliography}
